\documentclass[10pt,twocolumn,letterpaper]{article}

\usepackage{cvpr}              % To produce the CAMERA-READY version
\usepackage{times}
\usepackage{epsfig}
\usepackage{graphicx}
\usepackage{amsmath}
\usepackage{amssymb}
\usepackage{multirow}
\usepackage{color}
\usepackage{bbm}
\usepackage{colortbl}
\definecolor{mygray}{gray}{.9}

\usepackage[dvipsnames]{xcolor}

\usepackage[pagebackref,breaklinks,colorlinks]{hyperref}

\def\paperID{20} % *** Enter the Paper ID here
\def\confName{3DV\xspace}
\def\confYear{2024\xspace}

\title{Dynamic Prototype Adaptation with Distillation for \\ Few-shot Point Cloud Segmentation}

\author{Jie Liu$^{1}$~, Wenzhe Yin$^{1}$, Haochen Wang$^{1}$~, Yunlu Chen$^2$, Jan-Jakob Sonke$^{3}$, Efstratios Gavves$^1$\\
{\normalsize{$^1$\{j.liu5, w.yin, h.wang3, E.Gavves\}@uva.nl}}~~~
{\normalsize{$^3$yunluche@andrew.cmu.edu}}~~~
{\normalsize{$^3$j.sonke@nki.nl}}\\
{\normalsize{$^1$University of Amsterdam, Netherlands}}~~~{\normalsize{$^2$Carnegie Mellon University, USA}}~~~
{\normalsize{$^3$The Netherlands Cancer Institute, Netherlands}}}

\begin{document}
\maketitle
\begin{abstract}
\label{abstract}
Few-shot point cloud segmentation seeks to generate per-point masks for previously unseen categories, using only a minimal set of annotated point clouds as reference. Existing prototype-based methods rely on support prototypes to guide the segmentation of query point clouds, but they encounter challenges when significant object variations exist between the support prototypes and query features. In this work, we present dynamic prototype adaptation (DPA), which explicitly learns task-specific prototypes for each query point cloud to tackle the object variation problem. DPA achieves the adaptation through prototype rectification, aligning 
vanilla prototypes from support with the query feature distribution, and prototype-to-query attention, extracting task-specific context from query point clouds. Furthermore, we introduce a prototype distillation regularization term, enabling knowledge transfer between early-stage prototypes and their deeper counterparts during adaption. By iteratively applying these adaptations, we generate task-specific prototypes for accurate mask predictions on query point clouds. Extensive experiments on two popular benchmarks show that DPA surpasses state-of-the-art methods by a significant margin,  e.g., 7.43\% and 6.39\% under the 2-way 1-shot setting on S3DIS and ScanNet, respectively. Code is available at \hyperlink{https://github.com/jliu4ai/DPA}{https://github.com/jliu4ai/DPA}.
\end{abstract}    
\section{Introduction}
\label{sec:intro}
Point cloud semantic segmentation is a critical computer vision task with wide-ranging applications, such as autonomous driving, robotics, and augmented reality. It involves assigning semantic labels to individual points in a 3D point cloud, providing valuable insights into the scene's geometry and semantics. Although existing point cloud segmentation methods~\cite{qi2017pointnet, qi2017pointnet++, hu2020randla,ye20183d, li2022hybridcr,li2023less,jiang2021guided,zhang2021perturbed} have achieved remarkable performance within the supervised learning paradigm, they heavily rely on large annotated training datasets, which are both time-consuming and labor-intensive to create. Furthermore, these methods often encounter challenges when confronted with novel classes not present in the training data.

%%%%%%%%%%%%%%%%%%%%%%%%%%%%%%%%%%%%%%%%%%%%
\begin{figure}[!t]
\vspace{-5mm}
\centering
\includegraphics[width=\linewidth]{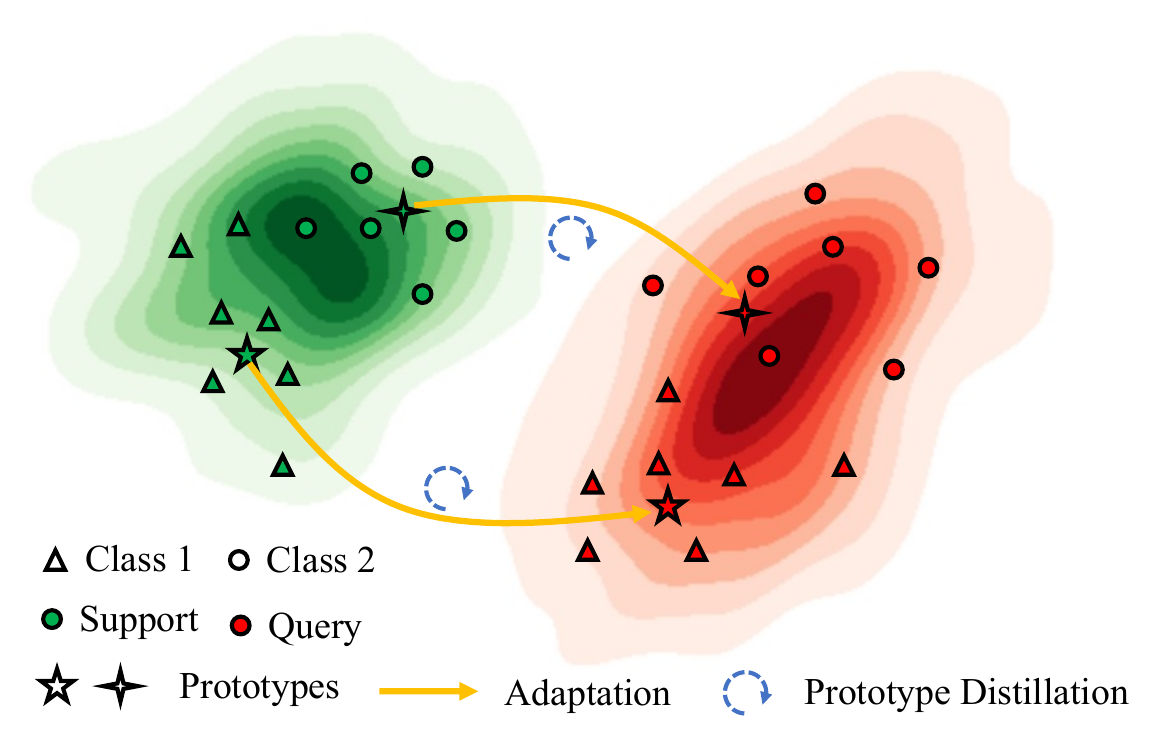}
\vspace{-6mm}
\caption{\textbf{Illustration of the proposed method.} Support and query point clouds often exhibit significant object variations, which can result in prototypes obtained from the support data being unsuitable for segmenting the query point cloud accurately. In this work, we propose to generate task-specific prototypes for the query point cloud by dynamically adapting prototypes with prototype distillation into query feature distribution.}
\label{fig:motivation}
\end{figure}

%%%%%%%%%%%%%%%%%%%%%%%%%%%%%%%%%%%%%%%%%%%%
By leveraging a small set of annotated support point clouds as a reference, few-shot point cloud segmentation methods~\cite{zhao2021few,mao2022bidirectional,zhang2023few,he2023prototype} provides a promising solution to address the aforementioned limitations. These methods follow a prototype-based paradigm, where prototypes are extracted from the support point clouds and utilized to segment the query point cloud. For instance, MPTI~\cite{zhao2021few} utilizes a transductive label propagation to exploit the affinity between multiple support prototypes and query features. BFG~\cite{mao2022bidirectional} introduces a bidirectional feature globalization strategy between labeled prototypes and query features to enhance the generalization ability of point features. Despite significant progress, existing methods in few-shot point cloud segmentation still encounter challenges due to notable object variations between support and query features~\cite{he2023prototype}. For example, objects from the same class but different scenes often exhibit different scales, appearances, and feature variations. Consequently, the vanilla prototypes generated from support features may not be well-suited for accurately segmenting query point clouds.

%%%%%%%%%%%%%%%%%%%%%%%%%%%%%%%%%%%%%%%%%%%%
To tackle the issue of object variations, we propose to generate task-specific prototypes for the query point cloud. This is achieved by dynamically adapting vanilla prototypes through  prototype distillation from the support to the query feature distribution. As depicted in Figure~\ref{fig:motivation}, there is typically a significant gap between the support and query feature distributions, directly applying vanilla prototypes to segment query point cloud would cause inferior performance. Our motivation is to map the vanilla prototypes from the support feature distribution to the query feature distribution, enabling the model to generate task-specific prototypes well-suited for accurately segmenting the query point cloud. The  prototype distillation process between the prototypes is introduced to enhance the prototype adaptation, enabling information exchange between prototypes from different adaptation stages.

%%%%%%%%%%%%%%%%%%%%%%%%%%%%%%%%%%%%%%%%%%%%
The prototype adaptation process consists of three key components: prototype rectification, prototype-to-query attention, and prototype distillation. Initially, vanilla prototypes are initialized using support features, which exhibit a significant feature gap with the query point cloud. To bridge this gap, prototype rectification is employed to map the vanilla prototypes from the support feature distribution to the query feature distribution, aligning the prototypes with the characteristics of the query point cloud. Additionally, the prototype-to-query attention mechanism aggregates context information from the query features into the prototypes, enabling the generation of task-specific prototypes for each query point cloud and enhancing the adaptation to specific segmentation tasks. To further improve prototype adaptation, a prototype distillation regularization term is proposed, enabling early-stage prototypes to learn from their deeper counterparts. This facilitates knowledge transfer and refines the prototype representation during adaptation. By iteratively applying the above adaptation process, we generate task-specific prototypes that act as optimal classifiers for the segmentation of query point clouds.

In a nutshell, the main contributions of our work are:

$\bullet$ We propose DPA, an end-to-end Dynamic Prototype Adaptation framework that addresses object variation issues between support and query features by adapting vanilla prototypes to task-specific prototypes.

$\bullet$ We introduce prototype rectification and prototype-to-query attention mechanisms to facilitate the adaptation process. Additionally, we design a prototype distillation regularization to further enhance the adaptation.

$\bullet$ We achieve state-of-the-art performance on two few-shot point cloud segmentation benchmarks, \textit{i.e.}, S3DIS and ScanNet. Notably, our method outperforms previous SOTA by 7.43\% and 6.39\% under the 2-way 1-shot setting on S3DIS and ScanNet, respectively.

\section{Related Work}
\label{sec:rw}

\subsection{Point Cloud Semantic Segmentation}
Point cloud semantic segmentation~\cite{hu2020randla,ye20183d, li2022hybridcr, li2023less,jiang2021guided,zhang2021perturbed}, which seeks to allocate specific labels to each individual point within a given point cloud, has been extensively studied in the computer vision community. PointNet~\cite{qi2017pointnet} is the pioneering work that adopts an end-to-end symmetric MLP network to segment raw point clouds. Later on point-based methods have sprung up due to its efficiency and simplicity. Point-wise MLP methods employ shared MLP layers as basic blocks of the network to extract features, such as PointNet++~\cite{qi2017pointnet++}, ShellNet~\cite{zhang2019shellnet}, and PointSIFT~\cite{jiang2018pointsift}. Inspired by the 2D convolution on image, point convolution methods design point-cloud-based convolution operations, including KPConv~\cite{thomas2019kpconv}, Deformable-Filter~\cite{xiong2019deformable}, DPC~\cite{engelmann2019dilated}, and PointConv~\cite{wu2019pointconv}. To extract the spatial geometric features of points, graph-based methods~\cite{wang2019dynamic, wang2019graph, landrieu2018large} construct graphs inside point sets and design novel graph convolutions. While these methods have demonstrated promising results in point cloud segmentation, they often cannot generalize well to novel categories not previously seen in the data. In this work, we tackle the point cloud semantic segmentation task of unseen classes with just a few labeled samples.

\subsection{Few-shot Point Cloud Semantic Segmentation}
Few-shot point cloud semantic segmentation extends the general point cloud semantic segmentation task to the few-shot scenario, where the model is endowed with the ability to segment novel classes with limited annotated support data. AttMPTI~\cite{zhao2021few} represents pioneering work in the realm of few-shot point cloud segmentation. It introduces a graph network to facilitate transductive inference for query point clouds. BFG~\cite{mao2022bidirectional} proposes a bidirectional feature globalization method to aggregate global information from both support and query data. SCAT~\cite{zhang2023few} introduces a class-specific attention based transformer network to improve both performance and efficiency for few-shot point cloud segmentation. QGPA~\cite{he2023prototype} proposes a prototype adaptation and projection mechanism to achieve prototype refinement, which leads to significant performance improvement. \cite{zhao2022crossmodal} introduces depth information to achieve cross-modal few-shot point cloud segmentation. Despite recent progress, the object variation problem, \textit{i.e.}, the objects with the same label but from different point clouds exhibit scale, appearance, and feature variations, is still challenging for current methods. In this work, we explore adopting dynamic prototype adaptation to address the object variation problem for few-shot point cloud segmentation.

\subsection{Knowledge Distillation} Knowledge distillation (KD)~\cite{buciluǎ2006model,hinton2015distilling} aims to transfer knowledge, such as logits or intermediate features, from a high-capacity teacher model to a lightweight student network. It has been applied in various domains, including computer vision and natural language processing, and has shown to be effective for model compression, regularization, and few-shot learning scenarios. Despite its competitive generalization improvement, pretraining great teacher model requires extra training time and computation cost~\cite{shen2022self}. To enhance efficiency in knowledge transferring and eliminate the requirement for extra teacher networks, self-distillation is proposed to distill knowledge from the model itself. There are three popular ways to achieve self knowledge distillation, \textit{i.e.}, 1) data-distortion based self-distillation \cite{lee2020self,xu2019data}, 2) regard history information as the teacher model, 3) distilling across auxiliary head~\cite{luan2019msd,zhang2019your}. In this work, we propose a prototype distillation scheme that regularizes the model to learn task-specific prototypes for the few-shot point cloud segmentation task.

\section{Method}
\label{sec:md}

%%%%%%%%%%%%%%%%%%%%%%%%%%%  model figure %%%%%%%%%%%%%%%%%%
\begin{figure*}[!t]
\vspace{-5mm}
\centering
\includegraphics[width=\textwidth]{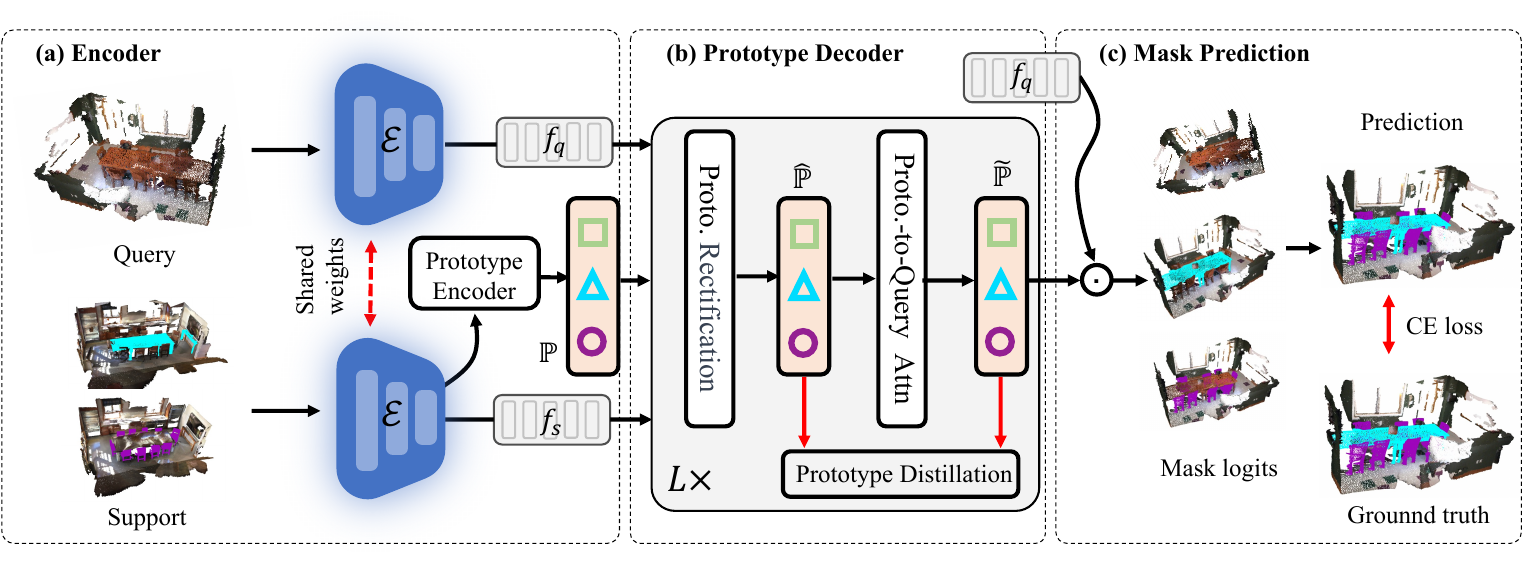}
\vspace{-6mm}
\caption{\textbf{Diagram of our proposed method.} \textbf{(a) Encoder:} Given the support and query point clouds, the feature encoder $\mathcal{E}$ extracts per-point feature $f_s$ and $f_q$, respectively. The prototype encoder transforms support features into the set of vanilla prototypes $\mathbb{P}$. \textbf{(b) Prototype Decoder:} The prototype decoder adapts vanilla prototypes to task-specific prototypes $\Tilde{\mathbb{P}}$ through $L$ transformer-based decoder blocks, which is composed of prototype rectification, prototype-to-query attention, and prototype distillation. The prototype distillation is introduced to enable early-stage prototypes $\hat{\mathbb{P}}$  to glean insights from their deeper counterparts $\Tilde{\mathbb{P}}$. \textbf{(c) Mask Decoder:} The mask prediction module generates per-class mask logits and then produces a final prediction using a softmax.}
\label{fig:model}
\end{figure*}

%%%%%%%%%%%%%%%%%%%%%%%%%%%  model figure %%%%%%%%%%%%%%%%%%
\subsection{Preliminary}
Following previous few-shot point cloud segmentation methods~\cite{zhao2021few,he2023prototype, mao2022bidirectional}, we adopt the episodic paradigm. Specifically, all classes in the dataset are divided into seen class set $C_{\text{seen}}$ and unseen class set  $C_{\text{unseen}}$, and $C_{\text{seen}} \cap C_{\text{unseen}}=\varnothing$. Each few-shot task (a.k.a. an episode) $\mathcal{T}=\{\mathcal{S}, \mathcal{Q}\}$, where $\mathcal{S}$ and $\mathcal{Q}$ are the support and query set, respectively, is instantiated as an $N$-way $K$-shot point cloud segmentation task. The support set $\mathcal{S}=\{(\mathbf{I}_s^{1, k}, \mathbf{M}_s^{1,k}),\ldots, (\mathbf{I}_s^{N, k}, \mathbf{M}_s^{N,k})\}_{k=1}^K$ consists of $K$ annotated support point clouds $\mathbf{I}_s^{n, k}$ and its binary mask $\mathbf{M}_s^{n,k}\in\mathbb{R}^{T\times 1}$ for each of the $N$ unique classes. The query set  $\mathcal{Q}=\{(\mathbf{I}_q^i,\mathbf{M}_q^i)\}_{i=1}^H$ contains $H$ (usually $H=N$) pairs of query point cloud $\mathbf{I}_q^i$ and its corresponding ground-truth mask $\mathbf{M}_q^i\in\mathbb{R}^{T\times 1}$, which is only available at training. Normally, each point cloud $\mathbf{I}\in \mathbb{R}^{T\times(3+f_0)}$ contains $T$ points associated with the coordinate information $\in\mathbb{R}^3$ and an additional feature $\in\mathbb{R}^{f_0}$, \textit{e.g.}, color. Given an $N$-way $K$-shot task, few-shot point cloud segmentation aims to predict the label $\mathbf{\hat{M}}_q\in\mathbb{R}^{T\times(N+1)}$ for any query point cloud $\mathbf{I}_q$ based on the support set $\mathcal{S}$.

%%%%%%%%%%%%%%%%%%%%%%%%%%%%%%%%%%%%%%%%%%%%%%%%%%%%%%%%%%%%%%%%%
\subsection{Overview}
The diagram of our proposed method is illustrated in Fig.~\ref{fig:model}, which consists of three important modules, encoder, prototype decoder, and mask prediction. Given support and query point clouds, the feature encoder $\mathcal{E}$ extracts per-point features $\mathbf{f}_s$ and $\mathbf{f}_q$ from the support and query point clouds, respectively. The prototype encoder generates initial prototypes $\mathbf{p}$ from the support feature. Then, the prototype decoder, which is composed of prototype rectification, prototype-to-query attention, and FFN, is introduced to refine initial prototypes and adapt them into task-specific prototypes $\mathbf{p}^*$. To achieve effective prototype adaptation, we introduce prototype distillation regularization to enable early-stage prototypes to glean insights from their deeper counterparts during adaptation. Finally, in the mask prediction module, the cosine similarity is employed between the query feature $\mathbf{f}_q$ and the task-specific prototype $\mathbf{p}^*$ to produce mask logits for generating the final mask prediction $\mathbf{M}_q^*$. The cross-entropy loss between prediction $\mathbf{M}_q^*$ and ground truth $\mathbf{M}_q$ is used to supervise the model training.

%%%%%%%%%%%%%%%%%%%%%%%%%%%%%%%%%%%%%%%%%%%%%%%%%%%%%%%%%%%%%%%%%
\subsection{Encoder}
\noindent\textbf{Feature Encoder.}
We adopt DGCNN~\cite{wang2019dynamic}, a dynamic graph CNN architecture, as the backbone of the feature encoder. Following ~\citet{zhao2021few}, we apply a self-attention network (SAN) on the generated semantic feature to mine semantic correlation between points in the global context. Given a task, we have the query point cloud $\mathbf{I}_q$, the support point cloud $\mathbf{I}_s^{n, k}$ and its binary mask $\mathbf{M}_s^{n,k}$, where $n\in\{1, \ldots, N\}$ and $k\in\{1, \ldots, K\}$ are the $N$-way $K$-shot indexes, respectively. Formally, the extracted features are given by:
\begin{equation}
\begin{aligned}
   \mathbf{f}_q &= \mathcal{E}(\mathbf{f}_q) \in \mathbb{R}^{T\times d} \\
    \mathbf{f}_s^{n,k} &= \mathcal{E}(\mathbf{I}_s^{n,k}\odot\mathbf{M}_s^{n,k})\in \mathbb{R}^{T\times d},
\end{aligned}
\end{equation}
where $\mathcal{E}$ represents the feature encoder, $T$ and $d$ denote the number and channel of point features, respectively. 

%%%%%%%%%%%%%%%%%%%%%%%%%%%%%%%%%%%%%%%%%%%%%%%%%%%%%%%%%%%%%%%%%
\noindent\textbf{Prototype Encoder.}
In the prototype encoder, we generate a prototype for each target category by conducting masked average pooling over support features with corresponding masks. The prototype for category $n$ is obtained by:
\begin{equation}
    \textbf{p}^{n}=\frac{1}{K}\sum_{k}\frac{\sum_x\mathbf{f}_{s,x}^{n,k}\mathbb{I}(\mathbf{M}_{s,x}^{n,k}=n)}{\sum_x\mathbb{I}(\mathbf{M}_{s,x}^{n,k}=n)},
\end{equation}
where $x\in{1,\ldots, N}$ denotes the coordinate position and $\mathbb{I}(*)$ is the indicator that outputs 1 when * is true. Besides, we compute a prototype $\mathbf{p}^0$ to represent the points that belong to background category:
\begin{equation}
    \textbf{p}^0=\frac{1}{NK}\sum_{n,k}\frac{\sum_x\mathbf{f}_{s,x}^{n,k}\mathbb{I}(\mathbf{M}_{s,x}^{n,k}\notin\{1,\ldots,N\})}{\sum_x\mathbb{I}(\mathbf{M}_{s,x}^{n,k}\notin\{1,\ldots,N\})}.
\end{equation}
Considering the $N$-way $K$-shot few-shot point cloud segmentation task, the prototype encoder generates the vanilla prototype set $\mathbb{P}=\{\mathbf{p}^0, \mathbf{p}^1, \ldots, \mathbf{p}^N\}$.

%%%%%%%%%%%%%%%%%%%%%%%%%%%%%%%%%%%%%%%%%%%%%%%%%%%%%%%%%%%%%%%%%
\subsection{Prototype Decoder}
Support and query point clouds often exhibit large object variations, which can result in the vanilla prototypes obtained from support objects being unsuitable for the segmentation of the query point cloud. To address this critical issue, we propose a transformer-based prototype decoder that adapts the vanilla prototypes to task-specific prototypes, ensuring their suitability for query point cloud segmentation. Specifically, the prototype decoder comprises of $L$ decoder blocks, with each block consisting of three key modules: prototype rectification, prototype-to-query attention, and prototype distillation. We take one block as an example and provide detailed explanations of these modules in the following text.

%%%%%%%%%%%%%%%%%%%%%%%%%%%%%%%%%%%%%%%%%%%%%%%%%%%%%%%%%%%%%%%%%
\noindent\textbf{Prototype Rectification.}
Due to the significant feature variations between the vanilla prototype set $\mathbb{P}$ and the query feature $\mathbf{f}_q$, we introduce the prototype rectification technique to adjust the vanilla prototypes and align them with the distribution of the query feature. Specifically, the rectified prototype $\mathbf{\hat{p}}^i\in \mathbb{R}^{1\times d}$, where $i\in \{0, 1, \ldots, N\}$, is defined as:
\begin{equation}
    \mathbf{\hat{p}}^i = \mathbf{p}^i+\mathbf{A}^i(\mathbf{p}^i)^\top,
\label{eq:pi_hat}
\end{equation}
where $\mathbf{A}^i\in\mathbb{R}^{d\times d}$ is the transformation matrix, and the original prototype is added as a residual connection for the stable training. The transformation matrix $\mathbf{A}^i$ plays a crucial role in achieving accurate prototype rectification. It needs to fulfill two essential conditions: 1) bridging the distribution gap between support and query features, and 2) introducing more diverse query information. To this end, we formulate the transformation matrix $\mathbf{A}^i$ as:
\begin{equation}
    \mathbf{A}^i = \text{softmax}((\mathbf{f}_q^i)^\top \mathbf{f}_s^*),
\label{eq:A}
\end{equation}
where $\mathbf{f}_s^*\in \mathbb{R}^{T\times d}$ is the support set representation that captures the statistics of all support features, and we simply define it as $\mathbf{f}_s^* =\frac{1}{NK}\sum_{n,k}\mathbf{f}_s^{n,k}$. The query feature $\mathbf{f}_q^i \in \mathbb{R}^{T\times d}$, $i\in \{0, 1, \ldots, N\}$ is from the query set that contains $N$ query point clouds, where $\mathbf{f}_q^0=\frac{1}{N}\sum_i\mathbf{f}_q^i$. The transformation matrix $\mathbf{A}^i$ establishes channel-wise correspondences between query and support features, thus bridging their distribution gap. Since the transformation matrix $\mathbf{A}^i$ varies with $\mathbf{f}_q^i$, we incorporate more diverse query information when computing the transformation matrix $\mathbf{A}^i$ for each prototype $\mathbf{p}^i$. Following Eq.(\ref{eq:pi_hat}) to Eq.(\ref{eq:A}), $N+1$ initial prototypes $\mathbb{P}=\{\mathbf{p}^0, \mathbf{p}^1, \ldots, \mathbf{p}^N\}$ are rectified as $\mathbb{\hat{P}}=\{\mathbf{\hat{p}}^0, \mathbf{\hat{p}}^1, \ldots, \mathbf{\hat{p}}^N\}$. The prototype rectification bridges the distribution gap between support and query features, enhancing compatibility between prototypes and the query point cloud.

%%%%%%%%%%%%%%%%%%%%%%%%%%%%%%%%%%%%%%%%%%%%%%%%%%%%%%%%%%%%%%%%%
\noindent\textbf{Prototype-to-Query Attention.} 
While prototype rectification aligns prototypes with query features, rectified prototypes still lack query-specific context information. To generate task-specific prototypes,  \textit{i.e.}, customized prototypes for a given query point cloud, we introduce prototype-to-query attention, which aggregates context information from the query feature into the prototypes.

Given the rectified prototype $\mathbf{\hat{p}}^i$ and a specific query feature $\mathbf{f}_q$, the cross-attention between the prototype and query feature is represented as:
\begin{equation}
    \mathbf{\dot{p}}^i = \text{softmax}(\mathbf{Q}\mathbf{K}^\top)\textbf{V} +\mathbf{\hat{p}}^i.
    \label{eq:cross}
\end{equation}
Here, $\mathbf{Q}=f_Q(\mathbf{\hat{p}}^i)\in\mathbb{R}^{1\times d}$, where $f_Q(\cdot)$ is the transformation function. $\mathbf{K},\mathbf{V}\in \mathbb{R}^{T\times d}$ are the query feature under transformation $f_K(\cdot)$ and $f_V(\cdot)$, respectively. We implement $f_Q, f_K, f_V$ as linear transformations.  We further feed $\mathbf{\dot{p}}^i$ into a feed-forward network (FFN) to get the final task-specific prototype:
\begin{equation}
    \mathbf{\Tilde{p}}^i = \text{FFN}(\mathbf{\dot{p}}^i)+\mathbf{\dot{p}}^i,
    \label{eq:ffn}
\end{equation}
where $\mathbf{\Tilde{p}}^i$ is the task-specific prototype tailored for the query point cloud, and the FFN is implemented as multiple 1D convolution layers. Following Eq.(\ref{eq:cross}) to Eq.(\ref{eq:ffn}), we could get the final task-specific prototype set $\mathbb{\Tilde{P}}=\{\mathbf{\Tilde{p}}^0, \mathbf{\Tilde{p}}^1, \ldots, \mathbf{\Tilde{p}}^N\}$, which is more accurate than vanilla prototype set $\mathbb{P}$ for the segmentation of query point cloud.

%%%%%%%%%%%%%%%%%%%%%%%%%%%%%%%%%%%%%%%%%%%%%%%%%%%%%%%%%%%%%%%%%
\noindent\textbf{Prototype Distillation.}
The proposed prototype-decoder adapts initial prototypes into task-specific prototypes. To enhance the prototype adaptation, we introduce a prototype distillation regularization term, which enables early-stage prototypes to glean insights from their deeper counterparts. Specifically, the prototype distillation regularization is formulated as a self-distillation paradigm :
\begin{equation}
    \mathcal{L}_{kd} = \text{KL}(\mathbb{\hat{P}}||\mathbb{\Tilde{P}}),
    \label{eq:kl}
\end{equation}
where KL refers to the Kullback-Leibler divergence. In our case, we utilize $\mathbb{\hat{P}}$ as the student prototypes and $\mathbb{\Tilde{P}}$ as the teacher prototypes. By distilling knowledge through the approximation of $\mathbb{\Tilde{P}}$ by $\mathbb{\hat{P}}$, the self-distillation process enables effective knowledge transfer between prototypes from different adaptation stages and enhances the adaptation of vanilla prototypes to task-specific prototypes.
%%%%%%%%%%%%%%%%%%%%%%%%%%%%%%%%%%%%%%%%%%%%%%%%%%%%%%%%%%%%%%%%%
\subsection{Mask Prediction}
After $L$ steps of prototype adaptation in the prototype decoder, we generate task-specific prototypes for query mask prediction. We adopt a non-parametric metric learning paradigm to perform segmentation. Since segmentation can be seen as classification at each point location, we calculate the distance between the query feature vector at each point location with each prototype. Subsequently, we apply a softmax function to the distances, generating mask logits that represent the different semantic classes, including the background. 
Concretely, given a distance function $d$, let $\mathbb{\Tilde{P}}=\{\mathbf{\Tilde{p}}^0, \mathbf{\Tilde{p}}^1, \ldots, \mathbf{\Tilde{p}}^N\}$ be the task-specific prototype set for the query feature $\mathbf{f}_q$. For each task-specific prototype $\Tilde{\mathbf{p}}^i$, mask logits are defined as:
\begin{equation}
    \mathbf{\hat{M}}_{q}^i = \frac{\text{exp}(-\alpha d(\mathbf{f}_{q},\Tilde{\mathbf{p}}^i))}{\sum_{\Tilde{\mathbf{p}}^i\in\mathbb{R}}\text{exp}(-\alpha d(\mathbf{f}_{q}, \Tilde{\mathbf{p}}^i))},
\end{equation}
where $\alpha$ is an amplification factor and we set it as 1.0 in our experiments. The predicted segmentation mask is then given by:
\begin{equation}
    \mathbf{\hat{M}}_q = \mathop{\text{arg max}}\limits_{i}\mathbf{\hat{M}}_{q}^i.
\end{equation}
%%%%%%%%%%%%%%%%%%%%%%%%%%%%%%%%%%%%%%%%%%%%%%%%%%%%%%%%%%%%%%%%%
\subsection{Objective}
At the training stage, the proposed model is supervised by two loss functions, \textit{i.e.}, the cross-entropy loss CE between the prediction $\mathbf{\hat{M}}_q$ and ground-truth $\mathbf{M}_q$, and the self-distillation regularization term. Formally, the overall loss is defined as:
\begin{equation}
    \mathcal{L} = \text{CE}(\mathbf{\hat{M}}_q, \mathbf{M}_q) +\gamma\mathcal{L}_{kd},
\end{equation}
Where $\gamma$ is the balancing loss weight and we set it as 0.1 in our experiments. The cross-entropy loss works as the main optimization objective, which aims to learn task-specific prototypes. The prototype distillation loss acts as a crucial regularization term, facilitating knowledge transfer from early-stage prototypes to their deeper counterparts. This process enhances the adaptation of vanilla prototypes into task-specific prototypes, resulting in improved results.

\section{Experiments}
\label{sec:exp}
\begin{table*}[htb]
    \centering
    \setlength{\abovecaptionskip}{0.cm}
    \setlength{\belowcaptionskip}{-0.cm}
    \scriptsize
    \begin{center}
    \caption{Performance on S3DIS dataset using mean-IoU metric (\%).  S$^{i}$ represents the split $i$ is used for model testing. The best results are masked in bold. Our method achieves consistent better performance than state-of-the-art methods across different $N$-way $K$-shot settings.} 
    \begin{tabular}{l|ccc|ccc|ccc|ccc}
    \toprule
    \label{table:s3dis}
    \multirow{3}*{\bf{Method}} 
          & \multicolumn{6}{c|}{\bf{2-way}}                                    & \multicolumn{6}{c}{\bf{3-way}}                            \\ \cline{2-13}
    ~   & \multicolumn{3}{c|}{\bf{1-shot}}    & \multicolumn{3}{c|}{\bf{5-shot}}  & \multicolumn{3}{c|}{\bf{1-shot}}  & \multicolumn{3}{c}{\bf{5-shot}}  \\ \cline{2-13}
    ~   & S$^{0}$ & S$^{1}$ & \bf{mean}       & S$^{0}$ & S$^{1}$ & \bf{mean}     & S$^{0}$ & S$^{1}$ & \bf{mean}     & S$^{0}$ & S$^{1}$ &\bf{mean}     \\
    \hline
   FT~\cite{zhao2021few}    & 36.34  & 38.79 & 37.57 & 56.49 & 56.99 & 56.74 & 30.05 & 32.19 & 31.12 & 46.88 & 47.57 & 47.23 \\
    ProtoNet~\cite{dong2018few}  & 48.39  & 49.98 & 49.19 & 57.34 & 63.22 & 60.28 & 40.81 & 45.07 & 42.94 & 49.05 & 53.42 & 51.24 \\
   MPTI~\cite{zhao2021few}  & 52.27  & 51.48 & 51.88 & 58.93 & 60.56 & 59.75 & 44.27 & 46.92 & 45.60 & 51.74 & 48.57 & 50.16 \\
   AttProtoNet~\cite{dong2018few}  & 50.98  & 51.90 & 51.44 & 61.02 & 65.25 & 63.14 & 42.16 & 46.76 & 44.46 & 52.20 & 56.20 & 54.20 \\
  AttMPTI~\cite{zhao2021few}  & 53.77  & 55.94 & 54.86 & 61.67 & 67.02 & 64.35 & 45.18 & 49.27 & 47.23 & 54.92 & 56.79 & 55.86 \\
  BFG~\cite{mao2022bidirectional}  & 55.60  & 55.98 & 55.79 & 63.71  &66.62 & 65.17 & 46.18 &48.36 & 47.27 &55.05 & 57.80  & 56.43 \\
  QGPA~\cite{he2023prototype}  & 59.45 & 66.08  & 62.76 & 65.40  & 70.30 & 67.85 & 48.99 & 56.57 & 52.78 & 61.27 & 60.81  & 61.04 \\
 \rowcolor[gray]{0.9} \textbf{Ours}  & \textbf{66.08}  & \textbf{74.30} & \textbf{70.19} & \textbf{71.10}  & \textbf{77.03} & \textbf{74.07} & \textbf{50.67} & \textbf{59.53} & \textbf{55.10} & \textbf{64.52} & \textbf{63.34}  &\bf{63.93} \\
 \bottomrule
\end{tabular}
\vspace{-3mm}
\end{center}
\end{table*}
\begin{table*}[htb]
   \setlength{\abovecaptionskip}{0.cm}
    \setlength{\belowcaptionskip}{-0.cm}
   \scriptsize
   \begin{center}
   \caption{Performance on ScanNet dataset using mean-IoU metric (\%).  S$^{i}$ represents the split $i$ is used for model testing. The best results are masked in bold. Our method achieves consistent better performance than state-of-the-art methods across different $N$-way $K$-shot settings.}
   \begin{tabular}{l|ccc|ccc|ccc|ccc}
   \toprule
   \label{table:scannet}
    \multirow{3}*{\bf{Method}} 
          & \multicolumn{6}{c|}{\bf{2-way}}                                    & \multicolumn{6}{c}{\bf{3-way}}                            \\ \cline{2-13}
    ~   & \multicolumn{3}{c|}{\bf{1-shot}}    & \multicolumn{3}{c|}{\bf{5-shot}}  & \multicolumn{3}{c|}{\bf{1-shot}}  & \multicolumn{3}{c}{\bf{5-shot}}  \\ \cline{2-13}
    ~   & S$^{0}$ & S$^{1}$ & \bf{mean}       & S$^{0}$ & S$^{1}$ & \bf{mean}     & S$^{0}$ & S$^{1}$ & \bf{mean}     & S$^{0}$ & S$^{1}$ &\bf{mean}     \\
    \hline
   FT~\cite{zhao2021few}      & 31.55  & 28.94 & 30.25 & 42.71 & 37.24 & 39.98 & 23.99 & 19.10 & 21.55 & 34.93 & 28.10 & 31.52  \\
   ProtoNet~\cite{dong2018few}   & 33.92  & 30.95 & 32.44 & 45.34 & 42.01 & 43.68 & 28.47 & 26.13 & 27.30 & 37.36 & 34.98 & 36.17  \\
   MPTI~\cite{zhao2021few}      & 39.27  & 36.14 & 37.71 &46.90 & 43.59 & 45.25 & 29.96 & 27.26 & 28.61 & 38.14 & 34.36 & 36.25  \\
   AttProtoNet~\cite{dong2018few} & 37.99  & 34.67 & 36.33 & 52.18 & 46.89 & 49.54 & 32.08 & 28.96 & 30.52 & 44.49 & 39.45 & 41.97  \\
   AttMPTI~\cite{zhao2021few}  & 42.55  & 40.83 & 41.69 & 54.00 & 50.32 & 52.16 & 35.23 & 30.72 & 32.98 & 46.74 & 40.80 & 43.77  \\
   BFG~\cite{mao2022bidirectional}  &42.15 &40.52 & 41.34 &51.23 &49.39 & 50.31 &34.12 &31.98 & 33.05 &46.25 &41.38 & 43.82 \\
     QGPA~\cite{he2023prototype}  & 57.08 & 55.94 & 56.51 & 64.55  & 59.64 & 62.10 & 55.27 & 55.60 & 55.44 & 59.02 & 53.16  & 56.09 \\
\rowcolor[gray]{0.9}   \textbf{Ours}  & \textbf{62.75} & \textbf{63.04} & \bf{62.90} & \textbf{67.19}  & \textbf{64.62} & \bf{65.91} & \textbf{61.97} & \textbf{61.72} & \bf{61.85} & \textbf{66.13} & \textbf{64.67}  &\bf{65.40} \\
\bottomrule
\end{tabular}
\vspace{-3mm}
\end{center}
\end{table*}
\begin{table}[!t]
\setlength{\abovecaptionskip}{0.cm}
\setlength{\belowcaptionskip}{-0.cm}
\begin{center}
\scriptsize
\caption{Ablation study of key components in the proposed method on S3DIS dataset under 2-way 1-shot setting. 'PR' denotes the prototype rectification, 'P2QA' denotes the prototype-to-query attention, and 'PD' denotes self-distillation regularization.}
\setlength{\tabcolsep}{4.4mm}{
\begin{tabular}{ccc|ccc}
\toprule
\label{ablation}
PR & P2QA & PD & $S^0$ & $S^1$ & mean \\ \hline
   &      &    &  50.45   &  51.12   &  50.79    \\
 \checkmark  &      &    & 60.42    & 68.55    & 64.49   \\ 
 \checkmark   &  \checkmark     &    & 64.35    &  70.72   &  67.54    \\
  \checkmark  &  \checkmark     &  \checkmark   & \textbf{66.08}    & \textbf{74.30}    &  \textbf{70.19}    \\\bottomrule
\end{tabular}}
\vspace{-5mm}
\end{center}
\end{table}

\subsection{Datasets and Evaluation Metrics}
\textbf{Datasets.}
The proposed method is evaluated on two point cloud segmentation benchmarks, \textit{i.e.}, Stanford Large-Scale 3D Indoor Spaces (\textbf{S3DIS})~\cite{armeni20163d} and \textbf{ScanNet}~\cite{dai2017scannet}. S3DIS is composed of 272 point cloud rooms in six indoor environments. The annotation for the point clouds has 12 semantic classes and one background class annotated as a cluster. ScanNet contains 1,513 point cloud scans from 707 special indoor scenes, which provides 20 semantic classes in addition to one background class annotated with annotated space. Compared with S3DIS, ScanNet includes more diverse room types, such as living rooms and bathrooms. 
%Following similar data pre-processing as in \cite{zhao2021few}, we 
Our data pre-processing follows \citet{zhao2021few} which
splits S3DIS and ScanNet into 7,547 and 36,359 blocks, respectively. From each block, 2,048 samples are randomly sampled. For each dataset, all semantic classes are divided into two disjoint class subsets. Then we perform two-fold cross-validation with one subset for training and the other for testing. We adopt a similar strategy as in \citet{zhao2021few} to sample multiple $N$-way $K$-shot episodes for model training and 100 episodes for model evaluation.

\noindent\textbf{Evaluate Metrics.}
We adopt the widely-used metric in point cloud semantic segmentation,\textit{i.e.}, mean Interaction over Union (mean-IoU), to evaluate our model. The mean-IoU is obtained by averaging over the set of testing classes. Considering different performance on each class split, we also report mean value of results from split-0 ($\text{S}^0$) and split-1 ($\text{S}^1$) for a comprehensive comparison.

\subsection{Implementation Details}
The implementation of our model is divided into three stages, \textit{i.e.}, backbone pretraining, meta-training, and meta-testing. For the pretraining stage, we adopt DGCNN~\cite{wang2019dynamic} as the feature extractor and additional MLP layers as the classifier. The pretraining is conducted on the base (seen) classes, we set the batchsize to 32 and learning rate is 0.001. We pretrain the model with Adam optimizer ($\beta_1=0.9$, $\beta_2=0.999$) for 150 epochs on both S3DIS and ScanNet. During the meta-training stage, we initialize the feature extract with pretrained weights and adopt Adam optimizer to update all parameters. The initial learning rate is set as 0.001 decays by half every 5,000 iterations. The hyper-parameters $\gamma$ and $L$ are set as 0.1 and 1 in all our experiments, respectively. We meta-train the model for 40,000 iterations, in which each episode is constructed based on randomly selected classes. During meta-testing, we randomly sample 100 episodes from unseen classes to perform model evaluation. All our models are implemented with PyTorch and trained on one NVIDIA 3090 GPU. 

\subsection{Ablation Study}
In this section, we perform an ablation study on the S3DIS dataset under the 2-way 1-shot setting to validate the effectiveness of our proposed DPA (Dynamic Prototype Adaptation) method. As a baseline model, we use ProtoNet, which directly employs support prototypes for segmentation without any adaptation.

\begin{figure}[!t]
\vspace{-5mm}
\centering
\includegraphics[width=\linewidth]{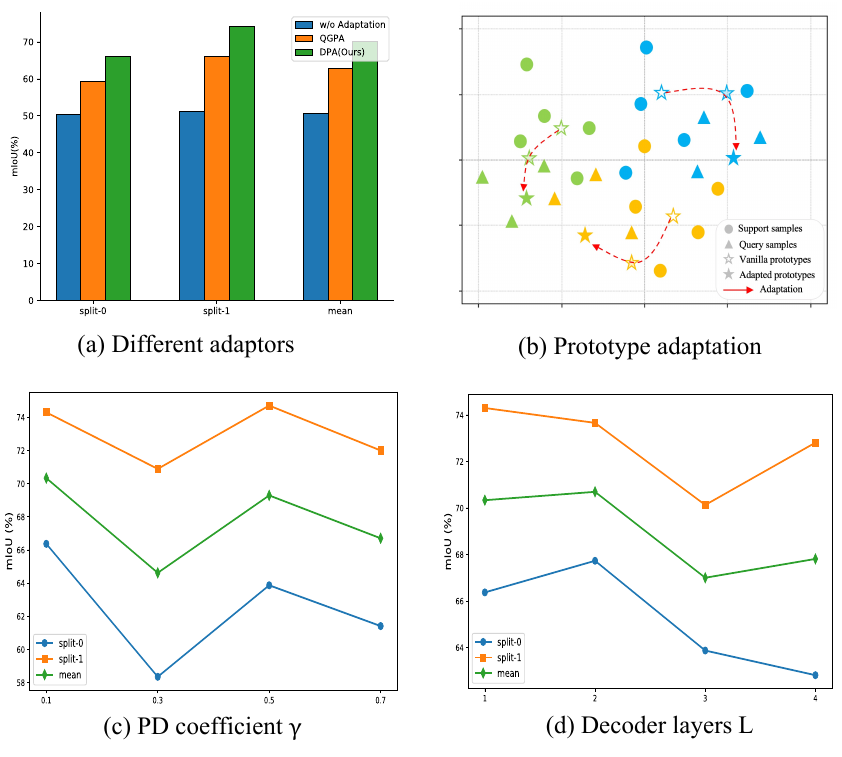}
\vspace{-6mm}
\caption{Ablation study of modules and hyper-parameters under 2-way 1-shot setting on S3DIS dataset. (a) Effects of different adaptors. (b)  t-SNE visualization of adaptation process on 3-way 5-shot task. (c) Effects of coefficient of prototype distillation (PD) loss $\gamma$. (d) Effects of number of decoder layers $L$.}
\label{fig:hyper}
\vspace{-3mm}
\end{figure}

\noindent\textbf{Effects of Prototype Adaptors.} In our study, we evaluate the effectiveness of prototype adaptation and compare our method with the previous prototype adaptor, QGPA~\cite{he2023prototype}, as well as the baseline model without any adaptation. As shown in Figure~\ref{fig:hyper} (a), both QGPA and our method DPA significantly outperform the baseline model. For instance, DPA surpasses the baseline by about 20\% in mean IoU. This indicates the crucial role of prototype adaptation in few-shot point cloud segmentation, especially when support and query point clouds exhibit large object variations.
Moreover, our DPA method consistently outperforms QGPA on both the S3DIS and ScanNet datasets, \textit{e.g.}, 70.19\% vs. 62.76 in mean. This demonstrates the superiority of our approach in achieving effective prototype adaptation and further improving segmentation performance.  We undertake a qualitative analysis of the
adaptation process of our DPA framework during the test time in Figure~\ref{fig:hyper} (b) using t-SNE~\cite{van2008visualizing}. The visualization underscores the importance of the prototype adaptation process in attaining improved performance in few-shot point cloud segmentation task.

\noindent\textbf{Benefits of Prototype Rectification (PR).}
As shown in Table~\ref{ablation}, the inclusion of prototype rectification in DPA significantly improves segmentation performance. For instance, it leads to a 13.70\% increase (64.49\% vs. 50.79\%) in mean IoU for the split $S^0$ and $S^1$, showcasing the effectiveness of prototype rectification in enhancing segmentation. With prototype rectification, DPA aligns initial prototypes from the support set to the query feature distribution, addressing the object variation issue and producing more accurate and task-specific prototypes for improved segmentation results.

%%%%%%%%%%%%%%%%%%%%%%%%%%%%%%%%%%%%%%%%%%%%
%%%%%%%%%%%%%%%%%%%%%%%%%%%%%%%%%%%%%%%%%%%

%%%%%%%%%%%%%%%%%%%%%%%%%%%  model figure %%%%%%%%%%%%%%%%%%

\noindent\textbf{Benefits of Prototype-to-Query Attention (P2QA).}
In Table~\ref{ablation}, the prototype-to-query attention further improves the model's performance by 3.05\% (67.54\% vs. 64.49\%). This improvement is attributed to the prototype-to-query attention mechanism, which aggregates task-specific context information from the query point cloud into the prototypes. As a result, the model generates more accurate and task-specific prototypes, leading to improved segmentation performance. The prototype-to-query attention enhances the model's ability to capture relevant information from the query point cloud, contributing to better segmentation.

\noindent\textbf{Benefits of Prototype Distillation (PD).}
In Table~\ref{ablation}, we demonstrate the effects of the prototype distillation term. As shown in the table, the prototype distillation term between prototype from prototype rectification (PR) and prototype-to-query attention (P2QA) results in a 2.65\% performance gain. This validates that self-distillation between prototypes effectively enables knowledge transfer and enhances the adaptation of initial prototypes to task-specific prototypes, leading to superior performance. The self-distillation mechanism plays a critical role in refining the prototypes and improving the segmentation results by facilitating information exchange between prototypes from different adaptation stages.

\noindent\textbf{Effects of Hyper-parameters.} In Figure~\ref{fig:hyper} (c) and (d), we illustrate the effects of hyperparameters $\gamma$ (coefficient of prototype distillation loss) and $L$ (number of decoder layers), respectively. As shown in Figure~\ref{fig:hyper} (c), we experiment with $\gamma$ values in the range [0.1, 0.3, 0.5, 0.7]. Larger values of $\gamma$ tend to cause unstable training, resulting in degraded performance. It is essential to find an appropriate balance for $\gamma$ to ensure stable and effective training. 
For the number of decoder layers $L$ (Figure~\ref{fig:hyper} (d)), we observe that increasing $L$ initially leads to improved performance. However, after a certain point, further increasing $L$ can cause overfitting and deteriorate the results. Therefore, selecting an optimal value for $L$ is crucial to achieving the best trade-off between model complexity and performance.

%%%%%%%%%%%%%%%%%%%%%%%%%%%  visualization figure %%%%%%%%%%%%%%%%%%
\begin{figure*}[!t]
\vspace{-5mm}
\centering
\includegraphics[width=\textwidth]{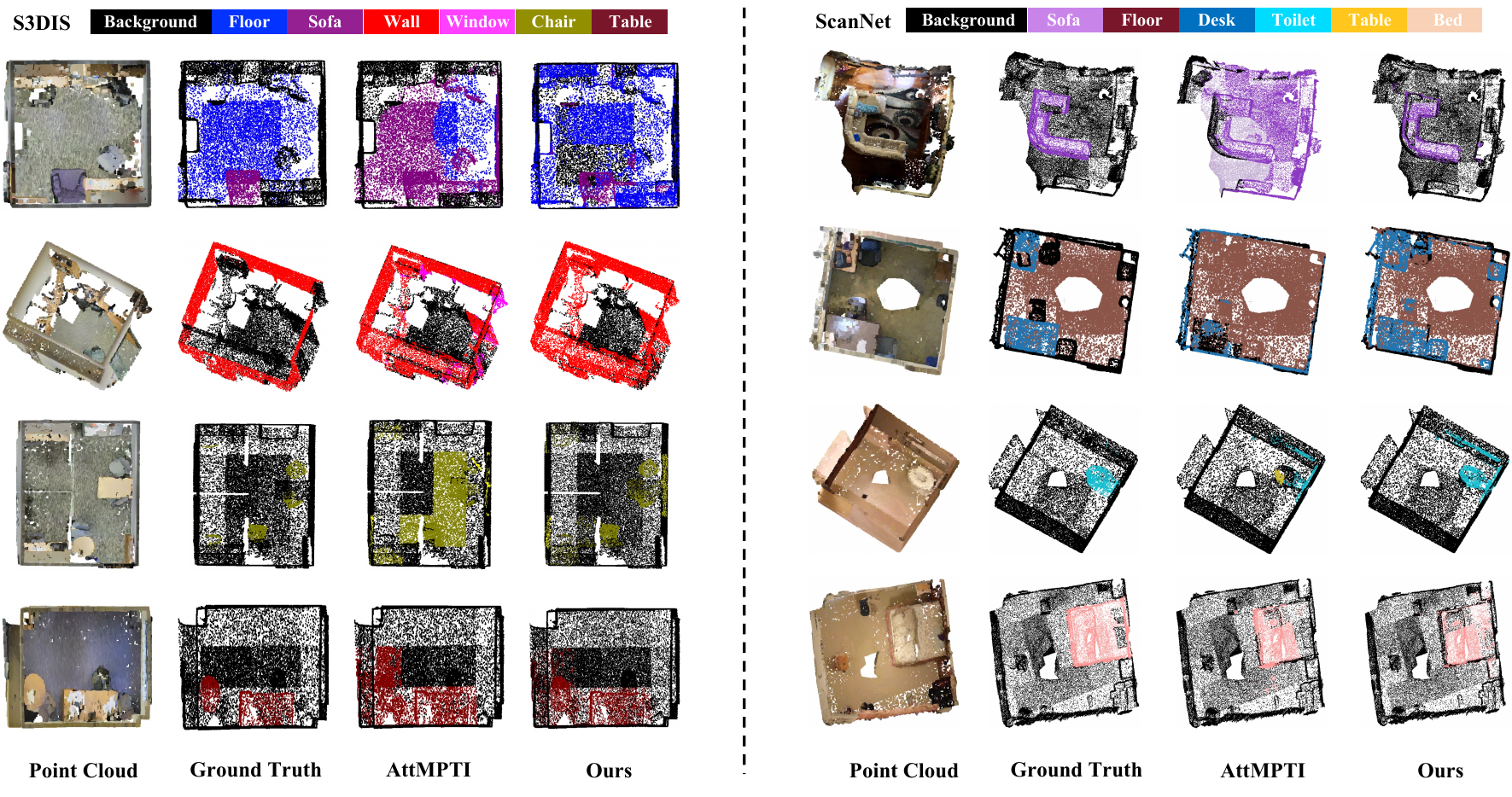}
\vspace{-6mm}
\caption{Quantitative results of our method in 2-way 1-shot point cloud semantic segmentation in comparison to Ground Truth and AttMPTI~\cite{zhao2021few}. Left: S3DIS. Right: ScanNet. Our method achieves consistently better segmentation results on both S3DIS and ScanNet.}
\label{fig:vis}
\vspace{-3mm}
\end{figure*}

%%%%%%%%%%%%%%%%%%%%%%%%%%%  visualization figure %%%%%%%%%%%%%%%%%%

\begin{table}[!t]
\setlength{\abovecaptionskip}{0.cm}
\setlength{\belowcaptionskip}{-0.cm}
\begin{center}
\scriptsize
\caption{Analysis of computational cost and experimental results under 2-way 1-shot setting. Our method achieves better balance between computation cost and experiment results.}
\label{table:cc}
\setlength{\tabcolsep}{3.2mm}{
\begin{tabular}{c|ccc|cc}
\toprule
Methods & Memory & \#Params & FPS & S3DIS & ScanNet \\ \hline
  attMPTI & 4.06G & 357.82K & 10.8 & 54.86 & 41.69 \\
  QGPA & 3.49G & 2.79M & 16.9 & 61.09 & 56.12 \\  
  DPA  & 4.21G  & 4.85M    & 15.3    & 70.19    & 62.90     \\\bottomrule
\end{tabular}}
\end{center}
\vspace{-5mm}
\end{table}

\subsection{Results and Analysis}

\textbf{Results on S3DIS.}
In Table~\ref{table:s3dis}, we present a comprehensive comparison of our proposed method with state-of-the-art (SOTA) methods, showcasing our quantitative results on the S3DIS dataset. Our proposed method consistently and significantly outperforms the compared methods across all four settings, including 2/3-way 1/5-shot scenarios. Notably, our method achieves a performance boost of 7.43\% (70.19\% vs. 62.76\%) and 6.22\% (74.07\% vs. 67.85\%) over the previous SOTA method QGPA~\cite{he2023prototype} in the 2-way 1-shot and 2-way 5-shot settings, respectively. Furthermore, our model demonstrates superior performance compared to QGPA by 2.32\% and 2.89\% in the 3-way 1-shot and 3-way 5-shot settings, respectively. Additionally, our model exhibits substantial improvement over the baseline model AttProtoNet~\cite{dong2018few}, for instance, with a notable margin of 15.1\% for the split 0 in the 2-way 1-shot setting.

\noindent\textbf{Results on ScanNet.}
Compared with S3DIS, ScanNet is a more challenging dataset, which contains more complex scenes. In Table~\ref{table:scannet}, we present the quantitative results on the ScanNet dataset. Our proposed method consistently and significantly outperforms the compared methods across all four settings. Specifically, when compared to the previous state-of-the-art method QGPA, our model achieves improvements of 6.39\% and 3.81\% in the 2-way 1-shot and 2-way 5-shot settings, respectively. These improvements are further extended to 6.41\% and 9.31\% in the 3-way 1-shot and 3-way 5-shot settings, respectively. Moreover, our method demonstrates significant performance gains over AttMPTI~\cite{zhao2021few} and BFG~\cite{mao2022bidirectional}, surpassing them by 21.21\% and 21.56\% in the 2-way setting, respectively. Overall, our proposed method achieves superior performance on the ScanNet dataset, showcasing its effectiveness and superiority compared to existing state-of-the-art methods.

\noindent\textbf{Qualitative Results.}
Figure~\ref{fig:vis} illustrates the qualitative results of our proposed method in 2-way 1-shot point cloud semantic segmentation on the S3DIS and ScanNet datasets. We compare our results with the ground truths and the predictions from AttMPTI. Across both datasets, our method consistently outperforms AttMPTI. For instance, in the first row of S3DIS examples, our method effectively distinguishes between the floor and the sofa, while AttMPTI mistakenly predicts some parts of the floor as the sofa. Similar results are observed in the first row of ScanNet examples. Furthermore, our method excels in segmenting small objects, such as chairs (the third row) in S3DIS and toilets (the third row) in ScanNet. The superior performance of our method stems from the task-specific prototypes, which are tailored to each specific query point cloud by dynamic prototype adaptation with prototype distillation, leading to improved segmentation performance.

\noindent\textbf{Computational Complexity.}
In Table~\ref{table:cc}, we present the number of parameters and computational complexity of our proposed method, along with a comparison to the previous state-of-the-art (SOTA) methods, AttMPTI~\cite{zhao2021few} and QGPA~\cite{he2023prototype}.
Our method has more parameters, primarily due to the prototype decoder, compared to AttMPTI and QGPA. However, despite the increase in parameters, our model achieves significantly better performance. For example, on the S3DIS dataset, our model outperforms AttMPTI and QGPA by 15.33\% and 9.10\%, respectively. Additionally, our model achieves comparable Frames Per Second (FPS) with QGPA (15.3 vs 16.9), which is significantly better than attMPTI. Our proposed model strikes a good balance between model performance and computational complexity, offering superior segmentation results with reasonable computational efficiency.

\noindent\textbf{Limitations.} While our method consistently outperforms previous methods, we have observed some challenges in distinguishing semantic-similar objects. For example, in the second row of ScanNet in Figure~\ref{fig:vis}, our method mistakenly recognizes chairs as desks. To address this issue, future work could explore incorporating multi-scale information to capture long-range context. Additionally, introducing semantic information, such as class name representations, as auxiliary constraints could also enhance the model's ability to handle semantic-similar objects and achieve more accurate segmentation results. 

\section{Conclusion}
\label{sec:con}
We propose dynamic prototype adaptation (DPA) for few-shot point cloud segmentation, addressing feature variations between support and query point clouds through effective prototype adaptation. DPA consists of three key components, prototype rectification, prototype-to-query attention, and prototype distillation. Prototype rectification aligns initial prototypes from the support with the query feature distribution, adapting them to the query point cloud. Prototype-to-query attention aggregates context information from the query feature into prototypes, generating task-specific prototypes for each point cloud. To further enhance the adaptation process, we propose prototype distillation as a regularization technique that refines the prototypes by promoting information exchange between different adaptation stages. Extensive evaluations on the S3DIS and ScanNet datasets demonstrate that DPA achieves state-of-the-art performance, significantly surpassing previous methods.

{\small
\bibliographystyle{ieeenat_fullname}
\bibliography{main}

\begin{thebibliography}{33}
\providecommand{\natexlab}[1]{#1}
\providecommand{\url}[1]{\texttt{#1}}
\expandafter\ifx\csname urlstyle\endcsname\relax
  \providecommand{\doi}[1]{doi: #1}\else
  \providecommand{\doi}{doi: \begingroup \urlstyle{rm}\Url}\fi

\bibitem[Armeni et~al.(2016)Armeni, Sener, Zamir, Jiang, Brilakis, Fischer, and
  Savarese]{armeni20163d}
Iro Armeni, Ozan Sener, Amir~R Zamir, Helen Jiang, Ioannis Brilakis, Martin
  Fischer, and Silvio Savarese.
\newblock 3d semantic parsing of large-scale indoor spaces.
\newblock In \emph{IEEE CVPR}, pages 1534--1543, 2016.

\bibitem[Buciluǎ et~al.(2006)Buciluǎ, Caruana, and
  Niculescu-Mizil]{buciluǎ2006model}
Cristian Buciluǎ, Rich Caruana, and Alexandru Niculescu-Mizil.
\newblock Model compression.
\newblock In \emph{Proceedings of the 12th ACM SIGKDD international conference
  on Knowledge discovery and data mining}, pages 535--541, 2006.

\bibitem[Dai et~al.(2017)Dai, Chang, Savva, Halber, Funkhouser, and
  Nie{\ss}ner]{dai2017scannet}
Angela Dai, Angel~X Chang, Manolis Savva, Maciej Halber, Thomas Funkhouser, and
  Matthias Nie{\ss}ner.
\newblock Scannet: Richly-annotated 3d reconstructions of indoor scenes.
\newblock In \emph{IEEE CVPR}, pages 5828--5839, 2017.

\bibitem[Dong and Xing(2018)]{dong2018few}
Nanqing Dong and Eric~P Xing.
\newblock Few-shot semantic segmentation with prototype learning.
\newblock In \emph{BMVC}, 2018.

\bibitem[Engelmann et~al.(2019)Engelmann, Kontogianni, and
  Leibe]{engelmann2019dilated}
Francis Engelmann, Theodora Kontogianni, and Bastian Leibe.
\newblock Dilated point convolutions: On the receptive field of point
  convolutions.
\newblock \emph{arXiv preprint arXiv:1907.12046}, 2, 2019.

\bibitem[He et~al.(2023)He, Jiang, Jiang, and Ding]{he2023prototype}
Shuting He, Xudong Jiang, Wei Jiang, and Henghui Ding.
\newblock Prototype adaption and projection for few-and zero-shot 3d point
  cloud semantic segmentation.
\newblock \emph{IEEE Transactions on Image Processing}, 2023.

\bibitem[Hinton et~al.(2015)Hinton, Vinyals, and Dean]{hinton2015distilling}
Geoffrey Hinton, Oriol Vinyals, and Jeff Dean.
\newblock Distilling the knowledge in a neural network.
\newblock \emph{arXiv preprint arXiv:1503.02531}, 2015.

\bibitem[Hu et~al.(2020)Hu, Yang, Xie, Rosa, Guo, Wang, Trigoni, and
  Markham]{hu2020randla}
Qingyong Hu, Bo Yang, Linhai Xie, Stefano Rosa, Yulan Guo, Zhihua Wang, Niki
  Trigoni, and Andrew Markham.
\newblock Randla-net: Efficient semantic segmentation of large-scale point
  clouds.
\newblock In \emph{IEEE CVPR}, pages 11108--11117, 2020.

\bibitem[Jiang et~al.(2021)Jiang, Shi, Tian, Lai, Liu, Fu, and
  Jia]{jiang2021guided}
Li Jiang, Shaoshuai Shi, Zhuotao Tian, Xin Lai, Shu Liu, Chi-Wing Fu, and Jiaya
  Jia.
\newblock Guided point contrastive learning for semi-supervised point cloud
  semantic segmentation.
\newblock In \emph{Proceedings of the IEEE/CVF international conference on
  computer vision}, pages 6423--6432, 2021.

\bibitem[Jiang et~al.(2018)Jiang, Wu, Zhao, Zhao, and Lu]{jiang2018pointsift}
Mingyang Jiang, Yiran Wu, Tianqi Zhao, Zelin Zhao, and Cewu Lu.
\newblock Pointsift: A sift-like network module for 3d point cloud semantic
  segmentation.
\newblock \emph{arXiv preprint arXiv:1807.00652}, 2018.

\bibitem[Landrieu and Simonovsky(2018)]{landrieu2018large}
Loic Landrieu and Martin Simonovsky.
\newblock Large-scale point cloud semantic segmentation with superpoint graphs.
\newblock In \emph{IEEE CVPR}, pages 4558--4567, 2018.

\bibitem[Lee et~al.(2020)Lee, Hwang, and Shin]{lee2020self}
Hankook Lee, Sung~Ju Hwang, and Jinwoo Shin.
\newblock Self-supervised label augmentation via input transformations.
\newblock In \emph{International Conference on Machine Learning}, pages
  5714--5724. PMLR, 2020.

\bibitem[Li et~al.(2023)Li, Shum, and Breckon]{li2023less}
Li Li, Hubert~PH Shum, and Toby~P Breckon.
\newblock Less is more: Reducing task and model complexity for 3d point cloud
  semantic segmentation.
\newblock In \emph{Proceedings of the IEEE/CVF Conference on Computer Vision
  and Pattern Recognition}, pages 9361--9371, 2023.

\bibitem[Li et~al.(2022)Li, Xie, Shen, Ke, Qiao, Ren, Lin, and
  Ma]{li2022hybridcr}
Mengtian Li, Yuan Xie, Yunhang Shen, Bo Ke, Ruizhi Qiao, Bo Ren, Shaohui Lin,
  and Lizhuang Ma.
\newblock Hybridcr: Weakly-supervised 3d point cloud semantic segmentation via
  hybrid contrastive regularization.
\newblock In \emph{Proceedings of the IEEE/CVF Conference on Computer Vision
  and Pattern Recognition}, pages 14930--14939, 2022.

\bibitem[Luan et~al.(2019)Luan, Zhao, Yang, and Dai]{luan2019msd}
Yunteng Luan, Hanyu Zhao, Zhi Yang, and Yafei Dai.
\newblock Msd: Multi-self-distillation learning via multi-classifiers within
  deep neural networks.
\newblock \emph{arXiv preprint arXiv:1911.09418}, 2019.

\bibitem[Mao et~al.(2022)Mao, Guo, Xiaonan, Yuan, and
  Guo]{mao2022bidirectional}
Yongqiang Mao, Zonghao Guo, LU Xiaonan, Zhiqiang Yuan, and Haowen Guo.
\newblock Bidirectional feature globalization for few-shot semantic
  segmentation of 3d point cloud scenes.
\newblock In \emph{2022 International Conference on 3D Vision (3DV)}, pages
  505--514. IEEE, 2022.

\bibitem[Qi et~al.(2017{\natexlab{a}})Qi, Su, Mo, and Guibas]{qi2017pointnet}
Charles~R Qi, Hao Su, Kaichun Mo, and Leonidas~J Guibas.
\newblock Pointnet: Deep learning on point sets for 3d classification and
  segmentation.
\newblock In \emph{IEEE CVPR}, pages 652--660, 2017{\natexlab{a}}.

\bibitem[Qi et~al.(2017{\natexlab{b}})Qi, Yi, Su, and Guibas]{qi2017pointnet++}
Charles~R Qi, Li Yi, Hao Su, and Leonidas~J Guibas.
\newblock Pointnet++: Deep hierarchical feature learning on point sets in a
  metric space.
\newblock \emph{arXiv preprint arXiv:1706.02413}, 2017{\natexlab{b}}.

\bibitem[Shen et~al.(2022)Shen, Xu, Yang, Li, and Guo]{shen2022self}
Yiqing Shen, Liwu Xu, Yuzhe Yang, Yaqian Li, and Yandong Guo.
\newblock Self-distillation from the last mini-batch for consistency
  regularization.
\newblock In \emph{Proceedings of the IEEE/CVF Conference on Computer Vision
  and Pattern Recognition}, pages 11943--11952, 2022.

\bibitem[Thomas et~al.(2019)Thomas, Qi, Deschaud, Marcotegui, Goulette, and
  Guibas]{thomas2019kpconv}
Hugues Thomas, Charles~R Qi, Jean-Emmanuel Deschaud, Beatriz Marcotegui,
  Fran{\c{c}}ois Goulette, and Leonidas~J Guibas.
\newblock Kpconv: Flexible and deformable convolution for point clouds.
\newblock In \emph{IEEE ICCV}, pages 6411--6420, 2019.

\bibitem[Van~der Maaten and Hinton(2008)]{van2008visualizing}
Laurens Van~der Maaten and Geoffrey Hinton.
\newblock Visualizing data using t-sne.
\newblock \emph{Journal of machine learning research}, 9\penalty0 (11), 2008.

\bibitem[Wang et~al.(2019{\natexlab{a}})Wang, Huang, Hou, Zhang, and
  Shan]{wang2019graph}
Lei Wang, Yuchun Huang, Yaolin Hou, Shenman Zhang, and Jie Shan.
\newblock Graph attention convolution for point cloud semantic segmentation.
\newblock In \emph{IEEE CVPR}, pages 10296--10305, 2019{\natexlab{a}}.

\bibitem[Wang et~al.(2019{\natexlab{b}})Wang, Sun, Liu, Sarma, Bronstein, and
  Solomon]{wang2019dynamic}
Yue Wang, Yongbin Sun, Ziwei Liu, Sanjay~E Sarma, Michael~M Bronstein, and
  Justin~M Solomon.
\newblock Dynamic graph cnn for learning on point clouds.
\newblock \emph{TOG}, 38\penalty0 (5):\penalty0 1--12, 2019{\natexlab{b}}.

\bibitem[Wu et~al.(2019)Wu, Qi, and Fuxin]{wu2019pointconv}
Wenxuan Wu, Zhongang Qi, and Li Fuxin.
\newblock Pointconv: Deep convolutional networks on 3d point clouds.
\newblock In \emph{IEEE CVPR}, pages 9621--9630, 2019.

\bibitem[Xiong et~al.(2019)Xiong, Ren, Liao, Wong, and
  Urtasun]{xiong2019deformable}
Yuwen Xiong, Mengye Ren, Renjie Liao, Kelvin Wong, and Raquel Urtasun.
\newblock Deformable filter convolution for point cloud reasoning.
\newblock \emph{arXiv preprint arXiv:1907.13079}, 2019.

\bibitem[Xu and Liu(2019)]{xu2019data}
Ting-Bing Xu and Cheng-Lin Liu.
\newblock Data-distortion guided self-distillation for deep neural networks.
\newblock In \emph{Proceedings of the AAAI Conference on Artificial
  Intelligence}, pages 5565--5572, 2019.

\bibitem[Ye et~al.(2018)Ye, Li, Huang, Du, and Zhang]{ye20183d}
Xiaoqing Ye, Jiamao Li, Hexiao Huang, Liang Du, and Xiaolin Zhang.
\newblock 3d recurrent neural networks with context fusion for point cloud
  semantic segmentation.
\newblock In \emph{ECCV}, pages 403--417, 2018.

\bibitem[Zhang et~al.(2023)Zhang, Wu, Wu, Zhao, and Wang]{zhang2023few}
Canyu Zhang, Zhenyao Wu, Xinyi Wu, Ziyu Zhao, and Song Wang.
\newblock Few-shot 3d point cloud semantic segmentation via stratified
  class-specific attention based transformer network.
\newblock \emph{arXiv preprint arXiv:2303.15654}, 2023.

\bibitem[Zhang et~al.(2019{\natexlab{a}})Zhang, Song, Gao, Chen, Bao, and
  Ma]{zhang2019your}
Linfeng Zhang, Jiebo Song, Anni Gao, Jingwei Chen, Chenglong Bao, and Kaisheng
  Ma.
\newblock Be your own teacher: Improve the performance of convolutional neural
  networks via self distillation.
\newblock In \emph{Proceedings of the IEEE/CVF International Conference on
  Computer Vision}, pages 3713--3722, 2019{\natexlab{a}}.

\bibitem[Zhang et~al.(2021)Zhang, Qu, Xie, Li, Zheng, and
  Li]{zhang2021perturbed}
Yachao Zhang, Yanyun Qu, Yuan Xie, Zonghao Li, Shanshan Zheng, and Cuihua Li.
\newblock Perturbed self-distillation: Weakly supervised large-scale point
  cloud semantic segmentation.
\newblock In \emph{Proceedings of the IEEE/CVF International Conference on
  Computer Vision}, pages 15520--15528, 2021.

\bibitem[Zhang et~al.(2019{\natexlab{b}})Zhang, Hua, and
  Yeung]{zhang2019shellnet}
Zhiyuan Zhang, Binh-Son Hua, and Sai-Kit Yeung.
\newblock Shellnet: Efficient point cloud convolutional neural networks using
  concentric shells statistics.
\newblock In \emph{Proceedings of the IEEE/CVF international conference on
  computer vision}, pages 1607--1616, 2019{\natexlab{b}}.

\bibitem[Zhao et~al.(2021)Zhao, Chua, and Lee]{zhao2021few}
Na Zhao, Tat-Seng Chua, and Gim~Hee Lee.
\newblock Few-shot 3d point cloud semantic segmentation.
\newblock In \emph{IEEE CVPR}, pages 8873--8882, 2021.

\bibitem[Zhao et~al.(2022)Zhao, Wu, Wu, Zhang, and Wang]{zhao2022crossmodal}
Ziyu Zhao, Zhenyao Wu, Xinyi Wu, Canyu Zhang, and Song Wang.
\newblock Crossmodal few-shot 3d point cloud semantic segmentation.
\newblock In \emph{Proceedings of the 30th ACM International Conference on
  Multimedia}, pages 4760--4768, 2022.

\end{thebibliography}
}

\end{document}